\documentclass[11pt]{article}
\makeatletter
\newlength\titlebox
\twocolumn
\def\addcontentsline#1#2#3{}

\def\maketitle{%
  \par%
  \begingroup%
     \def\thefootnote{\fnsymbol{footnote}}%
     \def\@makefnmark{\rlap{$^{\@thefnmark}$\hss}}%
     \long\def\@makefntext##1{%
                  \parindent 1em\noindent%
                  \hbox to 1em{$^{\@thefnmark}$}##1}
     \twocolumn[\@maketitle] \@thanks%
  \endgroup%
  \setcounter{footnote}{0}%
  \let\maketitle\relax\let\@maketitle\relax%
  \gdef\@thanks{}\gdef\@author{}\gdef\@title{}%
  \let\thanks\relax}

\def\@maketitle{%
  \vbox to \titlebox{%
    \hsize\textwidth\linewidth\hsize%
    \vskip 0.125in minus 0.05in%
    \centering{\Large\bf \@title \par}%
    \vskip 0.2in plus 0.1fil minus 0.1in
    {\def\and{\unskip\enspace{\rm and}\enspace}%
     \def\And{\end{tabular}\hss \egroup \hskip 1in plus 2fil 
              \hbox to 0pt\bgroup\hss \begin{tabular}[t]{c}\bf}%
     \def\AND{\end{tabular}\hss\egroup \hfil\hfil\egroup
              \vskip 0.25in plus 1fil minus 0.125in
              \hbox to \linewidth\bgroup\large \hfil\hfil
              \hbox to 0pt\bgroup\hss \begin{tabular}[t]{c}\bf}
    \hbox to \linewidth \bgroup\large \hfil\hfil
    \hbox to 0pt\bgroup\hss \begin{tabular}[t]{c}\bf\@author 
                            \end{tabular}\hss\egroup
    \hfil\hfil\egroup}
  \vskip 0.3in plus 2fil minus 0.1in
}}

\renewenvironment{abstract}{\section*{\centerline{Abstract}}}{}

\def\section{%
    \@startsection{section}{1}{\z@}%
                  {-2.0ex plus -0.5ex minus -0.3ex}%
                  {0.8ex plus 0.3ex minus 0.2ex}%
                  {\large\bf\raggedright}}
\def\subsection{%
    \@startsection{subsection}{2}{\z@}%
                  {-1.4ex plus -0.4ex minus -0.2ex}%
                  {0.6ex plus 0.2ex minus 0.1ex}%
                  {\normalsize\bf\raggedright}}
\def\subsubsection{%
    \@startsection{subsubsection}{3}{\z@}%
                  {-0.8ex plus -0.3ex minus -0.1ex}%
                  {0.4ex plus 0.1ex minus 0.1ex}%
                  {\normalsize\bf\raggedright}}
\def\paragraph{%
    \@startsection{paragraph}{4}{\z@}%
                  {-0.8ex plus -0.3ex minus -0.1ex}%
                  {-1em}%
                  {\normalsize\bf}}
\def\subparagraph{%
    \@startsection{subparagraph}{5}{\parindent}%
                  {0.4ex plus 0.3ex minus 0.1ex}%
                  {-1em}%
                  {\normalsize\bf}}

\labelwidth\leftmargini\advance\labelwidth-\labelsep \labelsep 5pt

\ifcase\@ptsize%
    \renewcommand{\normalsize}{
        \@setsize\normalsize{11.3pt}\xpt\@xpt%
        \abovedisplayskip 10\p@\@plus2\p@\@minus5\p@%
        \abovedisplayshortskip\z@\@plus3\p@%
        \belowdisplayshortskip 4\p@\@plus3\p@\@minus3\p@%
        \belowdisplayskip\abovedisplayskip%
        \let\@listi\@listI}%
 \or%
    \renewcommand{\normalsize}{
        \@setsize\normalsize{12.6pt}\xipt\@xipt%
        \abovedisplayskip11\p@\@plus2\p@\@minus4\p@%
        \abovedisplayshortskip\z@\@plus3\p@%
        \belowdisplayshortskip5\p@\@plus3\p@\@minus2\p@%
        \belowdisplayskip\abovedisplayskip%
        \let\@listi\@listI}%
 \or%
    \renewcommand{\normalsize}{
        \@setsize\normalsize{13pt}\xiipt\@xiipt%
        \abovedisplayskip 11\p@ \@plus3\p@ \@minus5\p@%
        \abovedisplayshortskip \z@ \@plus3\p@%
        \belowdisplayshortskip 5\p@ \@plus3\p@ \@minus2\p@%
        \belowdisplayskip\abovedisplayskip%
        \let\@listi\@listI}%
 \fi    

\ifcase\@ptsize%
    \renewcommand{\small}{
        \@setsize\small{10.5pt}\ixpt\@ixpt%
        \abovedisplayskip 8\p@ \@plus3\p@ \@minus3\p@%
        \abovedisplayshortskip \z@ \@plus2\p@%
        \belowdisplayshortskip 3\p@ \@plus2\p@ \@minus2\p@%
        \belowdisplayskip\abovedisplayskip%
        \def\@listi{\leftmargin\leftmargini%
                    \topsep 3.5\p@ \@plus1.5\p@ \@minus1.5\p@%
                    \parsep 1.5\p@ \@plus\p@ \@minus\p@%
                    \itemsep \parsep}}%
 \or%
    \renewcommand{\small}{
        \@setsize\small{11.3pt}\xpt\@xpt%
        \abovedisplayskip 9\p@ \@plus2\p@ \@minus4\p@%
        \abovedisplayshortskip \z@ \@plus3\p@%
        \belowdisplayshortskip 5\p@ \@plus2.5\p@ \@minus2.5\p@%
        \belowdisplayskip\abovedisplayskip%
        \def\@listi{\leftmargin\leftmargini%
                    \topsep 5\p@ \@plus2\p@ \@minus2\p@%
                    \parsep 2\p@ \@plus2\p@ \@minus\p@%
                    \itemsep \parsep}}%
 \or%
    \renewcommand{\small}{
        \@setsize\small{12pt}\xipt\@xipt%
        \abovedisplayskip 9\p@ \@plus3\p@ \@minus4\p@%
        \abovedisplayshortskip \z@ \@plus3\p@%
        \belowdisplayshortskip 5\p@ \@plus2.5\p@ \@minus2\p@%
        \belowdisplayskip\abovedisplayskip%
        \def\@listi{\leftmargin\leftmargini%
                    \topsep 5.5\p@ \@plus2.5\p@ \@minus2.5\p@%
                    \parsep 4\p@ \@plus1.5\p@ \@minus\p@%
                    \itemsep \parsep}}%
 \fi

\ifcase\@ptsize
    \renewcommand{\footnotesize}{
        \@setsize\footnotesize{9.3pt}\viiipt\@viiipt%
        \abovedisplayskip 5\p@ \@plus2\p@ \@minus3\p@%
        \abovedisplayshortskip \z@ \@plus\p@%
        \belowdisplayshortskip 2.5\p@\@plus\p@\@minus2\p@%
        \belowdisplayskip\abovedisplayskip%
        \def\@listi{\leftmargin\leftmargini%
                    \topsep 2.5\p@ \@plus\p@ \@minus\p@%
                    \parsep 1.5\p@ \@plus\p@ \@minus\p@%
                    \itemsep \parsep}}%
 \or%
    \renewcommand{\footnotesize}{
        \@setsize\footnotesize{10.3pt}\ixpt\@ixpt%
        \abovedisplayskip 7\p@ \@plus2\p@ \@minus4\p@%
        \abovedisplayshortskip \z@ \@plus\p@%
        \belowdisplayshortskip 3\p@ \@plus2\p@ \@minus2\p@%
        \belowdisplayskip\abovedisplayskip%
        \def\@listi{\leftmargin\leftmargini%
                    \topsep 3\p@ \@plus2\p@ \@minus2\p@%
                    \parsep 2\p@ \@plus\p@ \@minus\p@%
                    \itemsep \parsep}}%
 \or%
    \renewcommand{\footnotesize}{
        \@setsize\footnotesize{11pt}\xpt\@xpt%
        \abovedisplayskip 9\p@ \@plus2\p@ \@minus4\p@%
        \abovedisplayshortskip \z@ \@plus3\p@%
        \belowdisplayshortskip 5\p@ \@plus3\p@ \@minus3\p@%
        \belowdisplayskip\abovedisplayskip%
        \def\@listi{\leftmargin\leftmargini%
                    \topsep 4.5\p@ \@plus2\p@ \@minus2\p@%
                    \parsep 3\p@ \@plus\p@ \@minus\p@%
                    \itemsep \parsep}}%
 \fi

\ifcase\@ptsize%
    \renewcommand{\large}{\@setsize\large{13pt}\xiipt\@xiipt}
 \or%
    \renewcommand{\large}{\@setsize\large{13pt}\xiipt\@xiipt}
 \or%
    \renewcommand{\large}{\@setsize\large{16pt}\xivpt\@xivpt}
 \fi

\ifcase\@ptsize%
    \renewcommand{\Large}{\@setsize\Large{16pt}\xivpt\@xivpt}
 \or%
    \renewcommand{\Large}{\@setsize\Large{16pt}\xivpt\@xivpt}
 \or%
    \renewcommand{\Large}{\@setsize\Large{16pt}\xivpt\@xivpt}
 \fi

\ifcase\@ptsize%
 \or%
 \or%
 \fi

\ifcase\@ptsize%
    \def\@listi{\leftmargin\leftmargini
                \topsep  6\p@ \@plus2\p@ \@minus2\p@%
                \parsep  2\p@ \@plus0.5\p@ \@minus\p@%
                \itemsep 2.5\p@ \@plus\p@ \@minus0.5\p@}%
 \or%
    \def\@listi{\leftmargin\leftmargini
                \topsep  8\p@ \@plus2\p@ \@minus2\p@%
                \parsep  3\p@ \@plus1.5\p@ \@minus\p@%
                \itemsep 3\p@ \@plus1.5\p@ \@minus\p@}%
 \or%
    \def\@listi{\leftmargin\leftmargini
                \topsep  9\p@ \@plus3\p@   \@minus4\p@%
                \parsep  4\p@  \@plus2\p@ \@minus\p@%
                \itemsep 4\p@  \@plus2\p@ \@minus\p@}%
 \fi
\let\@listI\@listi

\makeatother
\usepackage[dvips]{epsfig}
\usepackage{avm, chicago,amssymb,covingtn} 
\usepackage{vmargin}
\usepackage{times}
\setpapersize{A4}
\setmargins{2.2cm}{3.6cm}{16.5cm}{22.7cm}{0pt}{0pt}{0pt}{11mm}

\normalsize

\renewcommand{\r}[1]{(\ref{#1})}
\newcommand{\rr}[2]{(\ref{#1}#2)}

\avmhskip{.2em}
\avmvskip{.05ex}
\avmfont{\footnotesize\sf}
\avmsortfont{\footnotesize\it}
\avmvalfont{\footnotesize\it}
\avmoptions{center}

\newcommand{\feat}[1]{{\small\sf #1}}
\title{\vskip -4mm %
Processing Unknown Words in HPSG }
\author{Petra Barg \and Markus Walther\footnotemark \\
  Seminar f\"ur Allgemeine Sprachwissenschaft \\ Heinrich-Heine-Universit\"at D\"usseldorf \\
Universit\"atsstr. 1, D-40225 D\"usseldorf, Germany \\
\texttt{\{barg,walther\}@ling.uni-duesseldorf.de}}
\date{}

\begin{document}
\maketitle
\begin{abstract}
The lexical acquisition system presented in this paper
incrementally updates linguistic properties of unknown
words inferred from their surrounding context by parsing sentences 
with an HPSG grammar for German. We employ a gradual,
information-based concept of ``unknownness'' providing a uniform 
treatment for the range of completely known to maximally unknown 
lexical entries. ``Unknown'' information is viewed as revisable
information, which is either generalizable or specializable. 
Updating takes place after parsing, which only requires a modified
lexical lookup. Revisable pieces of information are identified by
grammar-specified declarations which provide access
paths into the parse feature structure. The updating mechanism revises
the corresponding places in the lexical feature structures iff the
context actually provides new information. For revising generalizable
information, type union is required. A worked-out example demonstrates the 
inferential capacity of our implemented system.
\end{abstract}
\section{Introduction}
\label{sec:introduction}
It is a remarkable fact that humans can often understand sentences
containing unknown words, infer their grammatical properties and
incrementally refine hypotheses about these words when encountering later instances.
In contrast, many current NLP systems still presuppose a complete
lexicon. Notable exceptions include \citeN{zernik:89},
\citeN{erbach:90}, \citeN{hastings.lytinen:94}. See \citeANP{zernik:89}
for an introduction to the general issues involved.

{\addtocounter{footnote}{1}\renewcommand{\thefootnote}{\fnsymbol{footnote}}
\footnotetext{This work was carried out within the {\em
 Sonderforschungsbereich 282 `Theorie des Lexikons'} (project B3),
funded by the German Federal Research  Agency DFG. 
We thank James Kilbury  and members of the B3 group for fruitful discussion.} \addtocounter{footnote}{-1}}
This paper describes 
an HPSG-based system which can incrementally
learn and refine properties of unknown words after
parsing individual sentences. It focusses on extracting linguistic
properties, as compared  to e.g. general concept learning
\cite{hahn.et.al:96}. 
Unlike \citeN{erbach:90}, however, it is not confined to simple morpho-syntactic
information but can also handle selectional restrictions,
semantic types and argument structure. Finally, while statistical
approaches like \citeN{brent:91} can gather e.g. valence information
from large corpora, we are more interested in full grammatical processing 
of individual sentences to maximally exploit each context.

The following three goals serve to structure our model. 
It should i) incorporate a {\em gradual}, information-based
conceptualization of ``unknownness''. Words are not 
unknown as a whole, but  may contain {\em unknown, i.e. revisable pieces of
  information}. Consequently, even known words can undergo revision
to e.g. acquire new senses. This view replaces the binary
distinction between open and closed class {\em words}. It should ii) maximally exploit the rich
representations and  modelling conventions of
HPSG and associated formalisms, with essentially the same grammar and lexicon
as compared to closed-lexicon approaches. This is important both to facilitate reuse
of existing grammars and to enable meaningful feedback for linguistic
theorizing.
Finally, it should iii) possess  domain-independent inference and
lexicon-updating capabilities. The grammar writer must be able to fully declare
which pieces of information are open to revision.

The system was implemented using MicroCUF, a simplified version of the CUF
typed unification formalism \cite{doerre.dorna:93} that we implemented in
SICStus Prolog. It shares both the feature logic and the definite
clause extensions with its big brother, 
but substitutes a closed-world type system for CUF's open-world
regime. A feature of our type system implementation that will be
significant later on is that type information in internal feature
structures (FSs) can be easily updated. 

The HPSG grammar developed with MicroCUF models a fragment of
German. Since our focus is on the lexicon, the range of syntactic
variation treated is currently limited to simplex sentences with
canonical word order. We have
incorporated some recent developments of HPSG, esp. the revisions of 
\citeN[ch. 9]{pollard.sag:94}, \citeN{manning.sag:95}'s proposal for an
independent level of argument structure and \citeN{bouma:97}'s use of
argument structure to eliminate procedural lexical rules in favour of
relational constraints. Our elaborate ontology of semantic types --
useful for non-trivial acquisition of selectional restrictions and
nominal sorts -- was derived from a systematic corpus study of a 
biological domain 
\cite[154-188]{knodel:80}. The grammar also covers all valence
classes encountered in the corpus. 
As for the lexicon format, we currently  list full forms only. 
Clearly, a morphology component would supply more contextual 
information from known 
affixes but would still require the processing of unknown stems.
\section{Incremental Lexical Acquisition}
\label{sec:incremental_lexical_acquisition}
When compared to a previous instance, a new sentential context can
supply either identical, more special, more general, or even
conflicting information along a given dimension. Example pairs
illustrating the latter three relationships are given under
\r{context.types2}-\r{context.types4} (words assumed to be unknown in bold face). 
\begin{examples}
\item \label{context.types2}
\begin{itemize}
\item[a.] 
Im {\bf Axon} tritt ein Ruhepotential auf. \\
 `a  rest potential occurs in the {\bf axon}'
\item[b.]
 Das Potential wandert \"uber das {\bf Axon}. \\
 `the potential travels along the {\bf axon}'
\end{itemize}
\item \label{context.types3}
\begin{itemize}
\item[a.] 
 Das Ohr {\bf reagiert} auf akustische Reize. \\
 `the ear {\bf reacts}  to acoustic stimuli'
\item[b.]
Ein Sinnesorgan {\bf reagiert} auf Reize. \\
 `a sense organ {\bf reacts} to stimuli'
\end{itemize}
\item \label{context.types4}
\begin{itemize}
\item[a.] 
Die Nase ist f\"ur Ger\"uche {\bf sensibel}. \\
 `the nose is {\bf sensitive} to smells' 
\item[b.]
Die {\bf sensible} Nase reagiert auf Ger\"uche. \\
 `the {\bf sensitive} nose reacts to smells'
\end{itemize}
\end{examples}
In contrast to \rr{context.types2}{a}, which provides the
information that the gender of {\em Axon} is not feminine (via
 {\em im}), the context
in \rr{context.types2}{b} is more specialized, assigning neuter
gender (via {\em das}). Conversely,
\rr{context.types3}{b} differs from \rr{context.types3}{a} in
providing a more general selectional restriction for the subject of
{\em reagiert}, since sense organs include ears as a
subtype. Finally, the adjective {\em sensibel} is used
predicatively in \rr{context.types4}{a}, but attributively in
\rr{context.types4}{b}. 
The usage types must be formally disjoint,
because some German adjectives allow for just
one usage  ({\em ehemalig} `former, attr.', {\em schuld} `guilty, pred.').

On the basis of contrasts like those in
\r{context.types2}-\r{context.types4} it makes sense to statically assign
 revisable information to one of two classes, namely 
{\em specializable} or {\em generalizable}.%
\footnote{The different behaviour underlying this classification has
  previously been noted by e.g. \citeN{erbach:90} and
  \citeN{hastings.lytinen:94} but received either no
  implementational status or no systematic association with arbitrary   {\em kinds of
 information}.%
}
Apart from the specializable kinds `semantic type of nouns' and `gender',
the inflectional class of nouns is
another candidate (given a morphological component). Generalizable kinds of information
 include `selectional restrictions of verbs and adjectives', `predicative vs attributive
usage of adjectives' as well as  `case and form of PP
arguments' and `valence class of verbs'. 
Note that specializable and generalizable information
can cooccur in a given lexical entry. A particular kind of
information may also figure in both classes, as
e.g. semantic type of nouns and selectional restrictions of verbs
are both drawn from the same semantic ontology.
Yet the former must be invariantly specialized -- independent of the
order in which contexts are processed --, whereas selectional
restrictions on NP complements should only become more general
with further contexts. 

\subsection{Representation}
\label{sec:representation}
\begin{figure}[htb]
    \begin{center}
      \epsfig{file=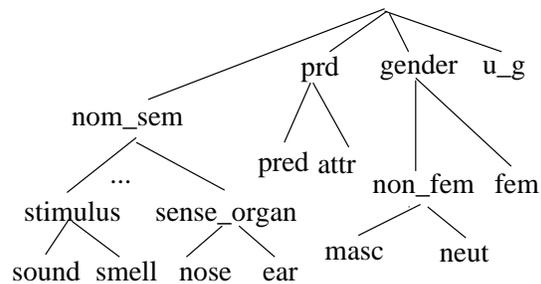}
      \caption{\it Excerpt from type hierarchy}
      \label{fig:typehierarchy}
    \end{center}
\vskip -0.95\baselineskip
  \end{figure}

We require all revisable or updateable information to be expressible as formal types.%
\footnote{In HPSG types are sometimes also referred to as sorts.}
As relational clauses can be defined to map types to
FSs, this is not much of a restriction in practice.
Figure \ref{fig:typehierarchy} shows a relevant fragment.
Whereas the combination of specializable information translates into simple
type unification (e.g. $non\_fem\wedge neut=neut$), combining
generalizable information requires {\em type 
union} (e.g. $pred\vee attr=prd$). The latter might pose problems for type systems requiring the
explicit definition of all possible unions, corresponding to least
common supertypes. However, type union is  easy for (Micro)CUF and
similar systems  which allow for arbitrary boolean combinations of types. 
Generalizable information exhibits another peculiarity: we need a disjoint
auxiliary type {\em u\_g} to correctly mark the initial unknown information state.
\footnote{Actually, the situation is more symmetrical, as we need a
  dual type {\em u\_s} to correctly mark ``unknown'' {\em
    specializable} information. This prevents incorrect updating of
  known information. However, {\em u\_s} is unnecessary for the examples
  presented below.}
This is because `content' types like {\em prd, pred, attr} are to be interpreted as
recording what contextual information was  encountered in the past. Thus, using
any of these to prespecify the initial value -- either as the
side-effect of a feature appropriateness declaration (e.g. {\em prd})
or through grammar-controlled specification (e.g. {\em pred, attr}) -- would be
wrong (cf. $prd_{initial}\vee attr=prd$, but
$u\_g_{initial}\vee attr=u\_g\vee attr$).

Generalizable information evokes another question: can we simply have types like
those in fig. \ref{fig:typehierarchy} within HPSG
signs and do in-place type union, just like type unification? 
The answer is no, for essentially two reasons. First, we still want to rule out
ungrammatical constructions through (type) unification failure of
coindexed values, so that generalizable types cannot {\em always} 
be combined by nonfailing type union (e.g. {\em *der sensible  Geruch} 
`the sensitive smell' must be ruled out via
$sense\_organ\wedge smell=\bot$). We would ideally like to order
all type unifications pertaining to a value before all unions, but
this violates the order independence of constraint solving.
Secondly, we already know that a given informational token can
{\em simultaneously} be generalizable and specializable, e.g. 
by being coindexed through HPSG's valence principle. However, 
simultaneous in-place union and unification is contradictory. 

To avoid these problems and keep the declarative monotonic setting, we 
employ two independent features \feat{gen}  and  \feat{ctxt}. 
 \feat{ctxt} is the repository of contextually unified
information, where conflicts result in ungrammaticality. \feat{gen}
holds generalizable information. 
Since all \feat{gen} values  contain  {\em u\_g} as a type disjunct,
they  are always unifiable and thus not restrictive during the parse.
To nevertheless get correct \feat{gen} values 
 we perform type union  {\em after} parsing,
 i.e. during lexicon update. We will see below how this works out.

The last representational issue is how to identify  revisable
information in (substructures of) the parse FS. 
For this purpose the grammar defines revisability clauses like the following:
\begin{examples} 
\item \label{filter.examples}
\begin{itemize}
\item[a.] generalizable(\avmbox{1}, \avmbox{2}) :=
\begin{avm}
\[synsem\|loc\|cat\|head & \[\it adj \\ prd & \[ gen & \@1 \\ ctxt & \@2
\] \] \]
\end{avm}
\item[b.] specializable(\avmbox{1}) :=
\begin{avm}
\[synsem\|loc & \[ cat\|head \; \it noun \\ 
                           cont\|ind\|gend & \@1
                        \]
\] 
\end{avm}
\end{itemize}
\end{examples} 
\subsection{Processing}
\label{sec:processing}
The first step in processing sentences with unknown or revisable words
consists of conventional parsing. Any HPSG-compatible parser may be
used, subject to the obvious requirement that
lexical lookup must not fail if a word's phonology is unknown.
A canonical entry for such unknown words is defined  
as the disjunction of maximally underspecified generic lexical entries for nouns, adjectives
and verbs.

The actual updating of lexical entries consists of
 four major steps. Step 1 projects the parse FS
derived from the whole sentence onto all participating word
tokens. This results in word FSs which are contextually
enriched (as compared to their original lexicon state) and
disambiguated  (choosing the compatible disjunct per parse solution if
the entry was disjunctive).
It then filters the set of word FSs by unification with the right-hand side
of revisability clauses like in \r{filter.examples}. 
The output of step 1 is a list of update candidates for those words
which were unifiable.

Step 2 determines concrete update values for each word: for each matching
{\em generalizable} clause we take the type union  of the \feat{gen}
value  of the old, lexical state of the word
($LexGen$) with the \feat{ctxt} value of its parse projection ($Ctxt$)
: $TU=LexGen\cup Ctxt$.  For each matching {\em
  specializable(Spec)} clause we take the parse value {\em Spec}.

Step 3 checks whether updating would make a difference
w.r.t. the original lexical entry of each word. The condition to be met
by generalizable information is that $TU \supsetneq LexGen$, for
specializable information we similarly require $Spec \subsetneq LexSpec$.

In step 4 the lexical entries of words surviving step 3 are actually
modified. We retract the old lexical entry, revise the entry
and re-assert it. For words never encountered before, revision must
obviously be preceded by making a copy of the  generic unknown
entry, but with the new word's phonology. Revision itself is the
destructive modification of type information according to the
values determined in step 2, at the places in a word FS pointed to 
by the revisability clauses. This is 
easy in MicroCUF, as types are implemented via
the {\em  attributed variable} mechanism 
of SICStus Prolog, which allows us to substitute 
the type in-place. In comparison, general updating
of Prolog-encoded FSs would typically require the traversal of large structures and
be dangerous if structure-sharing between substituted and unaffected parts
existed. Also note that we currently assume DNF-expanded 
entries, so that updates work on the contextually selected disjunct. 
This can be motivated by the advantages of working with presolved 
structures at run-time, avoiding description-level operations and 
incremental grammar recompilation.
\subsection{A Worked-Out Example}
\label{sec:a_worked_example}
We will illustrate how incremental lexical revision works by going
through the examples under \r{we:sentence1}-\r{we:sentence3}.
\begin{examples}
\item \label{we:sentence1}
         Die {\bf Nase} ist ein Sinnesorgan. \\
         `the {\bf nose} is a sense organ'
\item \label{we:sentence2}
         Das Ohr {\bf perzipiert}. \\
         `the ear {\bf perceives}'
\item \label{we:sentence3}
         Eine verschnupfte {\bf Nase perzipiert}  den Gestank. \\
         `a bunged up {\bf nose perceives} the stench'
\end{examples}
The relevant substructures corresponding to the lexical FSs of the
unknown noun and verb involved are depicted in fig. \ref{worked_example}.
The leading feature paths {\small\textsf{synsem$|$loc$|$cont}} for {\em Nase} and
{\small\textsf{synsem$|$loc$|$cat$|$arg-st}} for {\em perzipiert} have been omitted.

After parsing \r{we:sentence1} the gender of the unknown noun {\em Nase} is
instantiated to {\em fem} by agreement with the determiner {\em
  die}. As the {\em specializable} clause \rr{filter.examples}{b}
matches and the \feat{gend} parse value differs from its lexical value
{\em gender}, {\em gender} is updated to {\em fem}.
Furthermore, the object's semantic type has
percolated to the subject {\em Nase}. Since the object's {\em
  sense\_organ} type differs from generic  
initial {\em nom\_sem}, {\em Nase}'s \feat{ctxt} value is updated as well.
In place of the still nonexisting entry for {\em perzipiert}, we
have displayed the relevant part of the generic unknown verb entry.
\newbox\myavmbox
\avmfont{\scriptsize\sf}
\avmvalfont{\scriptsize\it}
\setbox\myavmbox=\hbox{%
\begin{tabular}{@{}
ll}
{\bf \rule{0mm}{4mm}Nase} & {\bf perzipiert} \\[2ex]
{\small\bf after \r{we:sentence1}} \\[1ex]
\begin{avm}
 \[ gend & \it fem \\
                                gen & \it u\_g  \\ 
                                ctxt & \it sense\_organ
                             \] 
\end{avm} &
\begin{avm}
\[gen & \it u\_g \\ ctxt & \it arg\_struc \\
                           
                         \]
\end{avm} \\
{} & {} \\
{\small\bf after \r{we:sentence2}} \\[1ex]
\begin{avm}
\[ gend & \it fem \\ gen & \it u\_g \\ ctxt & \it sense\_organ
\] 
\end{avm} &
\begin{avm}
\[gen &  \it u\_g$\vee$npnom \\ ctxt & \it arg\_struc \\
                            args & \< \[loc\|cont &
                                                                    \[
                                                                    gen & \it u\_g$\vee$ear
                                                                    \\
                                                                       ctxt & \it nom\_sem
                                                                    \]
                                           \]                    
                                           \|\underline{\phantom{x}} \> 
                         \]
\end{avm} \\
{\small\bf after \r{we:sentence3}} \\
\begin{avm}
\[ gend & \it fem \\ gen & \it u\_g \\ ctxt & \it nose\] 
\end{avm} &
\begin{avm}
\[gen &  \it u\_g$\vee$npnom$\vee$npnom\_npacc\\ ctxt & \it arg\_struc \\
                            args & \< \[loc\|cont &
                                                                    \[
                                                                    gen & \it u\_g$\vee$sense\_organ \\
                                                                       ctxt & \it nom\_sem
                                                                    \]
                                           \] {\bf ,} \\                     
                                           \[loc\|cont &
                                                                    \[
                                                                    gen & \it u\_g$\vee$smell \\
                                                                       ctxt & \it nom\_sem
                                                                    \]
                                            \]
                                           \|\underline{\phantom{x}} \> 
                         \]
\end{avm}
\end{tabular}}
\begin{figure}[t]
\begin{flushleft}
\box\myavmbox
\caption{\it Updates on lexical FSs}
    \label{worked_example}
\end{flushleft}
\vskip -4.5\baselineskip
\end{figure}

Having parsed  \r{we:sentence2} the system then knows that {\em perzipiert} can be
used intransitively with a nominative subject referring to ears.
Formally, an HPSG mapping principle was successful in mediating
between surface subject and complement lists and the argument list. 
Argument list instantiations are themselves related to corresponding types by a further mapping.
On the basis of this type classification of
argument structure patterns, the parse derived the \feat{ctxt} value {\em
  npnom}. Since \feat{gen} values are {\em generalizable}, this new value is
unioned with the old lexical \feat{gen} value. Note that 
\feat{ctxt} is properly unaffected. The 
first (subject) element on the \feat{args} list itself is targeted by another revisability
clause. This has the side-effect of further instantiating the underspecified
lexical FS. 
Since selectional restrictions on nominal subjects must become
more general with new contextual evidence, the union of {\em ear} and
the old value {\em u\_g} is indeed appropriate. 

Sentence \r{we:sentence3} first of all provides more specific evidence
about the semantic type of partially known {\em Nase} by way of
attributive modification through {\em verschnupfte}. The system
detects this through the difference between lexical \feat{ctxt} value
{\em sense\_organ} and the parse value {\em nose}, so that the
entry is {\em specialized} accordingly. Since the subject's
\feat{synsem} value is
coindexed with the first \feat{args} element, 
\avmfont{\footnotesize\sf}
\avmvalfont{\footnotesize\it}
\begin{avm} \[ctxt & \it nose\] \end{avm} 
simultaneously appears in the FS of {\em perzipiert}. However,
the revisability clause matching there is of class {\em generalizable},
so union takes place, yielding $ear\vee nose=sense\_organ$ (w.r.t. the
 simplified ontology of fig. \ref{fig:typehierarchy} used in this
 paper). An analogous match with the second element 
of \feat{args} identifies the necessary update to be the unioning-in of
 {\em smell}, the semantic type of {\em Ge\-stank}. Finally, the system has learned that
an accusative NP object can cooccur with {\em perzipiert}, so the
argument structure type of \feat{gen} receives another
update through union with {\em npnom\_npacc}.
\section{Discussion}
\label{sec:discussion}
The incremental lexical acquisition approach described above 
attains the goals stated earlier. It realizes a gradual, information-based
conceptualization of unknownness by providing updateable formal types
-- classified as either {\em generalizable} or {\em specializable} --
together with grammar-defined revisability clauses. 
It maximally exploits standard HPSG representations, 
requiring moderate rearrangements in grammars at best while keeping with
the standard assumptions of typed unification formalisms.  One noteworthy
demand, however, is the need for a type union operation. 
Parsing is conventional modulo a modified lexical lookup. The actual lexical 
revision is done in a domain-independent postprocessing step guided by
the revisability clauses.

Of course there are areas requiring further consideration. 
In contrast to humans, who seem to leap to conclusions based on
incomplete evidence, our approach employs a conservative form 
of generalization, taking the disjunction of actually observed values only. While
this has the advantage of not leading to overgeneralization, the 
requirement of having to encounter all subtypes in order to infer their 
common supertype is not realistic (sparse-data problem). 
In \r{worked_example} {\em sense\_organ} as the semantic type 
of the first argument of {\em perzipiert} is only acquired because
the simplified  hierarchy in fig. \ref{fig:typehierarchy} has {\em nose} 
and {\em ear} as its only subtypes. 
Here the work of \citeN{li.abe:95} who use the MDL
principle  to generalize over the slots of observed case frames might prove fruitful.

An important question is how to  administrate alternative parses 
and their update
hypotheses. In {\em Das Aktionspotential} {\bf\em erreicht} {\em den Dendriten} `the
action potential reaches the dendrite(s)', {\em Dendriten} is ambiguous
between acc.sg. and dat.pl., giving rise to two valence
hypotheses {\em npnom\_npacc} and {\em npnom\_npdat} for {\em erreicht}. Details remain
to be worked out on how to delay the choice between such alternative
hypotheses until further contexts provide enough information.

Another topic concerns  the treatment of  
`cooccurrence restrictions'. In fig.
\ref{worked_example} the system has {\em independently} generalized over
the selectional restrictions for subject and object, yet there are
clear cases where this overgenerates 
(e.g. *{\em Das Ohr perzipiert den Gestank} `the ear perceives the
stench'). An idea worth exploring is to have  
a partial, extensible list of type cooccurrences, which is traversed
by a recursive principle at parse time. 

A more general issue is the apparent antagonism between the desire to
have both sharp grammatical predictions and continuing openness to
contextual revision.
If  after parsing \r{we:sentence3} we transfer the
fact that smells are acceptable objects to {\em 
  perzipiert}  into the restricting \feat{ctxt} feature, a later usage
with an object of type {\em sound} fails.
The opposite case concerns newly acquired specializable values. 
If in a later context these are used to update a \feat{gen} value, the result 
may be too general. 
It is a topic of future research when to consider information  certain and
when to make revisable information restrictive.

\small\label{sec:references}

\end{document}